\documentclass[11pt, a4paper, logo, onecolumn]{googledeepmind}

\pdftrailerid{redacted}

\makeatletter
\renewcommand\bibentry[1]{\nocite{#1}{\frenchspacing\@nameuse{BR@r@#1\@extra@b@citeb}}}
\makeatother

\usepackage{longtable}
\usepackage{dsfont}
\usepackage{gdm-colors}
\usepackage{multirow}
\usepackage{graphicx}
\usepackage{rotating}
\usepackage{geometry}
\usepackage{xcolor}
\usepackage{placeins}
\usepackage{tabularx,tabulary}
\newcolumntype{Y}{>{\centering\arraybackslash}X}
\usepackage{colortbl}
\usepackage{hhline}
\usepackage{comment}
\usepackage{tcolorbox}
\usepackage{subcaption}

\definecolor{jhubluelight}{RGB}{104,172,229}
\definecolor{jhublue}{RGB}{0,45,114}

\newcommand{\insightbox}[1]{%
    \begin{tcolorbox}[colframe=jhublue!70, colback=jhubluelight!5, boxrule=1pt, arc=2mm]
        \includegraphics[width=0.4cm]{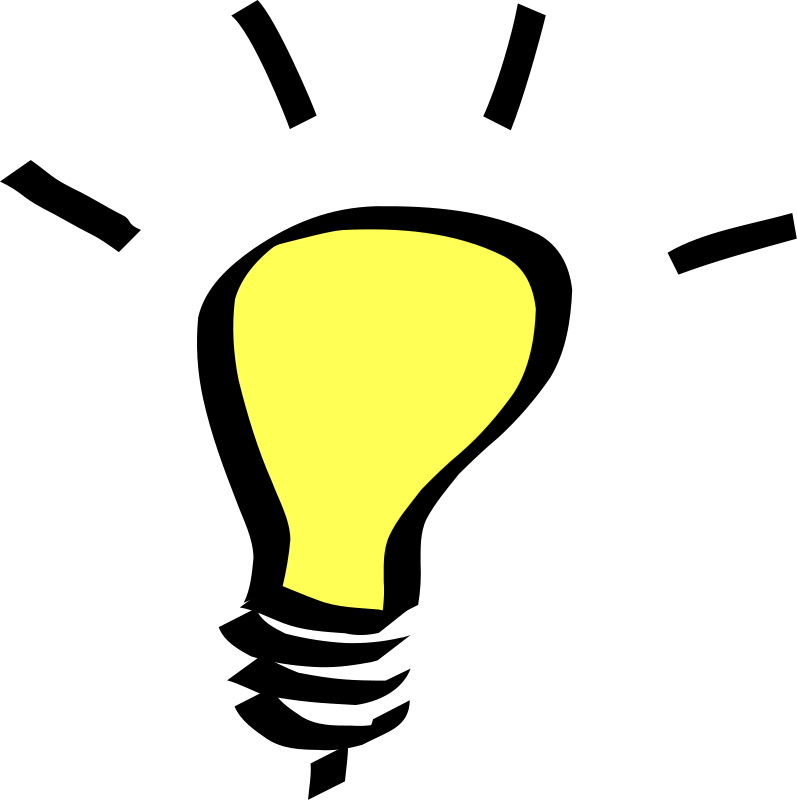}
        \textbf{\small#1}
    \end{tcolorbox}
}

\usepackage[numbers, sort&compress, square]{natbib}

\makeatletter
\DeclareRobustCommand\onedot{\futurelet\@let@token\@onedot}
\def\@onedot{\ifx\@let@token.\else.\null\fi}

\def\eg/{\emph{e.g}\onedot} \def\Eg/{\emph{E.g}\onedot}
\def\ie/{\emph{i.e}\onedot} \def\Ie/{\emph{I.e}\onedot}
\def\cf/{\emph{c.f}\onedot} \def\Cf/{\emph{C.f}\onedot}
\def\etc/{\emph{etc}\onedot} \def\vs/{\emph{vs}\onedot}
\def\wrt/{w.r.t\onedot} \def\dof/{d.o.f\onedot}
\def\etal/{\emph{et al}\onedot}

\newcommand{\arxiv}[1]{}
\makeatother

\newcolumntype{L}[1]{>{\raggedright\let\newline\\\arraybackslash\hspace{0pt}}m{#1}} 

\title{\modelname: Towards Explicit and Generalizable 3D Spatial Reasoning}
\newcommand{\modelname}{SpatialReasoner}

\newcommand{\titlerunning}{SpatialReasoner}

\renewcommand{\today}{April 2025}

\author[*]{Wufei Ma}
\author[*]{Yu-Cheng Chou}
\author[*]{Qihao Liu}
\author[ \hspace{-0.6ex}]{Xingrui Wang}
\author[$\dagger$]{Celso de Melo}
\author[$\circ$]{Jianwen Xie}
\author[ \hspace{-0.6ex}]{Alan Yuille}

\affil[ \hspace{-0.7ex}]{Johns Hopkins University}
\affil[*]{Equal contribution}
\affil[$\dagger$]{DEVCOM Army Research Laboratory}
\affil[$\circ$]{Lambda Inc}

\begin{abstract}
Despite recent advances on multi-modal models, 3D spatial reasoning remains a challenging task for state-of-the-art open-source and proprietary models.
Recent studies explore data-driven approaches and achieve enhanced spatial reasoning performance by fine-tuning models on 3D-related visual question-answering data.
However, these methods typically perform spatial reasoning in an implicit manner and often fail on questions that are trivial to humans, even with long chain-of-thought reasoning.
In this work, we introduce \modelname, a novel large vision-language model (LVLM) that addresses 3D spatial reasoning with explicit 3D representations shared between multiple stages--3D perception, computation, and reasoning.
Explicit 3D representations provide a coherent interface that supports advanced 3D spatial reasoning and improves the generalization ability to novel question types.
Furthermore, by analyzing the explicit 3D representations in multi-step reasoning traces of \modelname, we study the factual errors and identify key shortcomings of current LVLMs.
Results show that our \modelname{} achieves improved performance on a variety of spatial reasoning benchmarks, outperforming Gemini 2.0 by 9.2\% on 3DSRBench, and generalizes better when evaluating on novel 3D spatial reasoning questions.
Our study bridges the 3D parsing capabilities of prior visual foundation models with the powerful reasoning abilities of large language models, opening new directions for 3D spatial reasoning.

\vspace{.1in}
Project page (code, models and data): \href{https://spatial-reasoner.github.io}{spatial-reasoner.github.io}
\end{abstract}

\begin{document}

\maketitle

\section{Introduction} \label{sec:intro}

3D spatial reasoning studies how models perceive, understand, and reason about 3D object properties and spatial relationships.
It is not only a fundamental task for vision-language models to achieve human-level intelligence, but also crucial to a range of downstream applications in robotics~\cite{pgvlm2024,huang2024rekep} and embodied AI~\cite{cheng2024navila}.
Despite the recent advancements of large multi-modal models, such as GPT-4o and Qwen2.5-VL, their 3D spatial reasoning capabilities remain limited and fall far behind human-level performance~\cite{chen2024spatialvlm,ma20243dsrbench,yang2024thinking}.
Recent 3D-aware large vision-language models (LVLMs) injected 3D knowledge by fine-tuning the model on synthetic 3D-related question-answer pairs~\cite{chen2024spatialvlm,ma2025spatialllm} and achieved improved performance on spatial reasoning benchmarks~\cite{ma20243dsrbench,tong2024cambrian,yang2024thinking}.
On the other hand, large proprietary models such as Gemini 2.0~\cite{google2024gemini2} have advanced 3D parsing by directly predicting 3D object bounding boxes, enabling the development of powerful generalist robotics models~\cite{team2025geminirobotics}.

\begin{figure}[t]
    \centering
    \includegraphics[width=\linewidth]{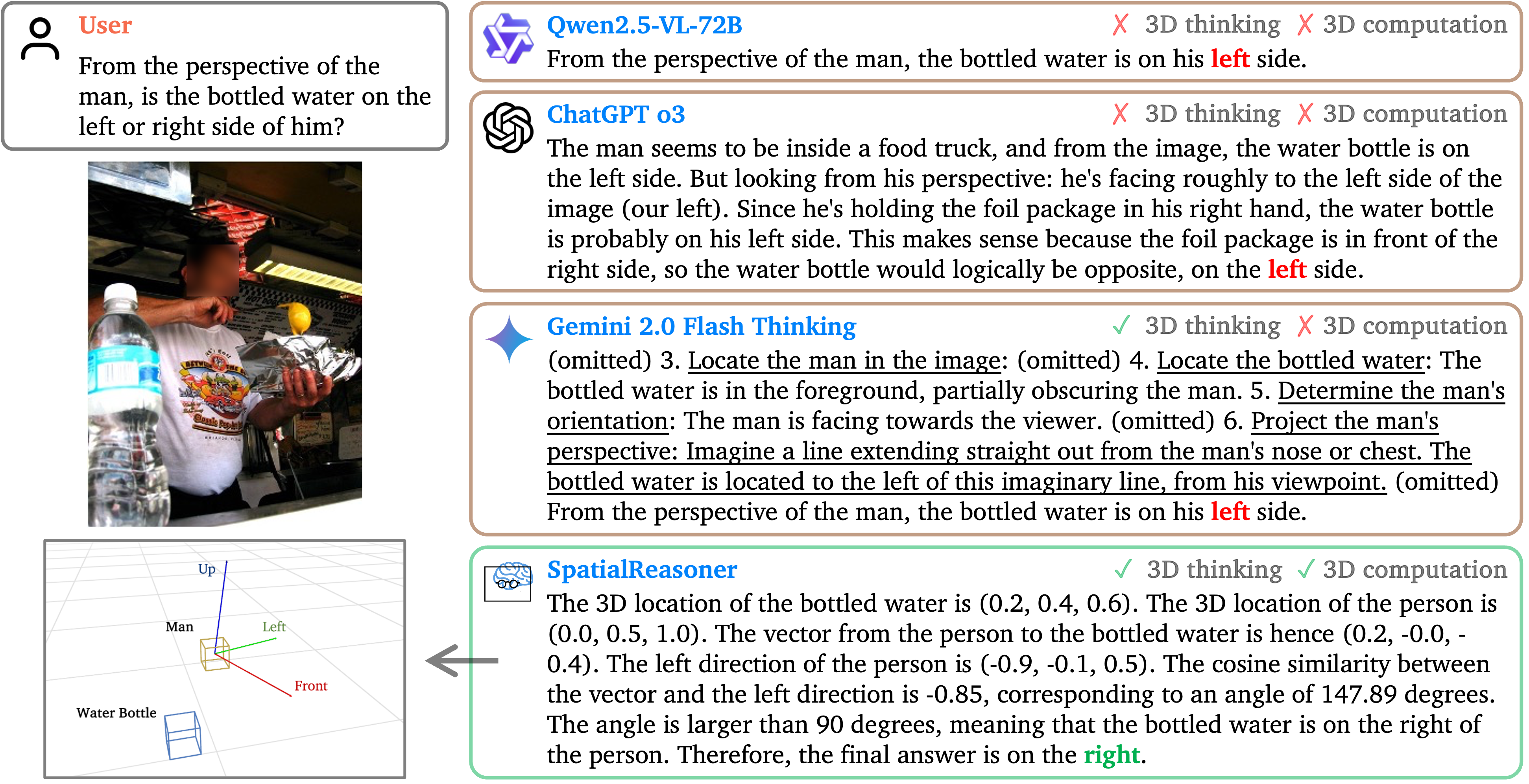}
    \caption{\textbf{Comparing 3D spatial reasoning of our SpatialReasoner with previous state-of-the-art models.} Our \modelname{} builds on explicit 3D representations, performs 3D computation, and reasons about the final answer. Although Gemini 2.0 can also break down complex 3D spatial reasoning questions into small and tractable steps, it lacks reliable 3D computation that leads to the correct answer.}
    \label{fig:teaser}
\end{figure}

%
We identify two key challenges in 3D spatial reasoning: (1) \textit{3D thinking}--the ability to decompose a complex 3D spatial reasoning question into small, manageable steps, and (2) \textit{3D computation}--the ability to solve these thinking steps in a consistent and accurate manner.
As shown in Figure~\ref{fig:teaser}, prior reasoning methods adopt long chain-of-thought reasoning to tackle the problem but do not have explicit 3D computation.
ChatGPT o3~\cite{openai2025o3} fails to adopt a systematic approach to solving the problem and instead relies on other visual cues to assist the reasoning (\textit{e.g.}, location of the foil package). In contrast, the Gemini 2.0 thinking model~\cite{google2024gemini2} employs an organized strategy to tackle the problem but ultimately fails to arrive at the correct answer due to a lack of reliable 3D computation.

In this work, we present \modelname{}, a novel large vision-language model (LVLM) built with (1) explicit 3D representations and (2) enhanced and generalizable 3D thinking. Specifically, our \modelname{} adopts explicit 3D representations, such as 3D locations and orientations, as an interface that enables coherent and reliable reasoning across multiple stages, \textit{i.e.}, 3D perception, computation, and reasoning.
On the other hand, we would like to learn enhanced 3D thinking capabilities that generalize to novel question types not seen during training.
Hence we explore a two-stage training strategy following prior works~\cite{guo2025deepseek}. In Stage I, we apply supervised fine-tuning (SFT) to equip the LVLM with explicit 3D representations, enhancing 3D perception and computation capabilities of the model. Then in Stage II, we leverage reinforcement learning (RL) to develop robust and generalizable 3D thinking built on explicit 3D representations.


Inspired by previous 3D pseudo-annotation pipelines in~\cite{chen2024spatialvlm,cheng2024spatialrgpt,ma2025spatialllm}, we synthesize basic 3D perception, \textit{i.e.}, detection and pose estimation, and 3D computation question-answering data interleaved with explicit 3D representations. We further generate standard spatial reasoning question-answer pairs with chain-of-thought reasoning that breaks down complex 3D spatial reasoning questions into multiple steps--3D perception, computation, and reasoning. Experimental results demonstrate that our \modelname{} with explicit 3D representations can significantly enhance 3D spatial reasoning abilities of LVLMs and generalize to novel question types.

Besides enhancing 3D spatial reasoning capabilities of LVLMs, reasoning with explicit 3D representations allows us to interpret the reasoning process and to study the failure modes of LVLMs.
We find that the accuracy of 3D perception lags significantly behind that of 3D computation, suggesting that most errors in downstream VQA tasks still stem from failures in 3D perception.
Moreover, by predicting key 3D information, such as 3D object locations and orientations, as intermediate results, our \modelname{} enables compositional reasoning for 3D spatial tasks~\cite{wang20233d}.
This not only allows our method to generalize better to novel spatial reasoning questions, but also makes it easily extensible to other tasks that build on our explicit 3D representations.

Besides improving spatial reasoning performance on a variety of benchmarks, we experiment on various LVLMs fine-tuned with different combinations of data and training methods to study the key factors toward improved 3D spatial reasoning.
Our empirical results lead to the following insights:
%
%
(1) When developing 3D-aware VLMs, SFT offers a more scalable approach than RL that requires high-quality 3D-aware data, which is often difficult to obtain. Our recipe of RL followed by SFT achieves the best overall performance;
(2) 3D-aware LVLMs fine-tuned with RL generalize better than SFT when tested on novel 3D spatial reasoning questions, echoing prior findings~\cite{chu2025sft};
(3) Standard LVLMs often exploit 2D reasoning as a shortcut to tackle 3D spatial reasoning problems, whereas our \modelname{} avoids the spurious correlations and always reasons with explicit 3D representations, achieving improved and robust performance on challenging real-world 3D spatial reasoning datasets.

In summary, our contributions are as follows: (1) We introduce \modelname{}, a novel LVLM that solves 3D spatial reasoning problems with a compositional approach
based on explicit 3D representations. (2) Reasoning with 3D representations allows us to interpret the reasoning process and to study the failure modes of LVLMs. (3) Our \modelname{} effectively improves the 3D spatial reasoning performance on a range of benchmarks and extensive experimental results provide valuable findings on data and training strategy designs for future development of 3D-aware LVLMs.

\begin{figure}[t]
    \centering
    \includegraphics[width=\linewidth]{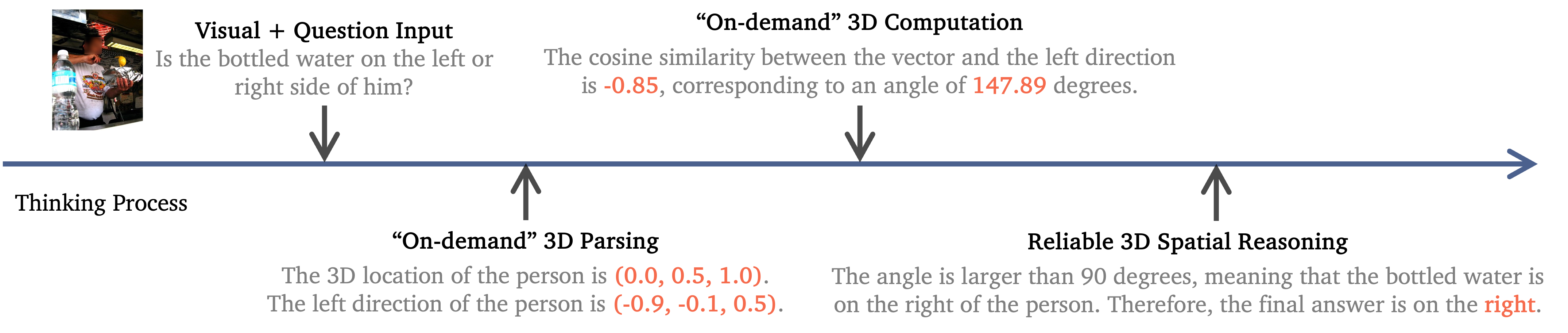}
    \caption{\textbf{Overview of our \modelname{} design.} Our \modelname{} adopts explicit 3D representations as an interface to enable coherent and reliable multi-stage spatial reasoning, \textit{i.e.}, 3D parsing, computation, and reasoning.}
    \label{fig:design}
\end{figure}

\section{Related Works} \label{sec:related}

\paragraph{3D Spatial reasoning.}
3D spatial reasoning explores how models perceive and reason about 3D object properties and relationships. Early works~\cite{wang20233d,wang2024compositional,wang2025pulsecheck457} studied this in simulated environments or datasets with object-level annotations~\cite{johnson2017clevr,li2023super,brazil2023omni3d,cheng2024spatialrgpt,tong2024cambrian,ma2025spatialllm}. More recent efforts construct benchmarks based on real-world imagery~\cite{ma20243dsrbench,yang2024thinking} and improve model performance via synthetic QA data~\cite{chen2024spatialvlm,cheng2024spatialrgpt,ma2025spatialllm,ray2024sat}. However, these models often rely on implicit reasoning, offering little interpretability or intermediate 3D computation. Moreover, it remains unclear whether the 3D spatial reasoning capabilities acquired from data-driven fine-tuning can generalize to novel question types that require complex 3D computation over different combinations of 3D perception outputs.

\paragraph{Explicit 3D representations.} Explicit 3D representations simplify reasoning and expose model failures. Simulation-based works~\cite{Mao2019NeuroSymbolic} used neural-symbolic methods to parse object-level structures, while PO3D~\cite{wang20233d} and its extensions~\cite{wang2024compositional,wang2025pulsecheck457} showed that structured visual modules enable interpretable reasoning~\cite{ma2022robust}. These results in simulation systems reveal the limitations of current LVLMs and highlight the importance of the visual module for a successful reasoning model.

\paragraph{Post-training.} Post-training aligns pre-trained models with downstream objectives via SFT~\cite{wei2021finetuned,zhou2023lima} and RL~\cite{bai2022training,bai2022constitutional,rafailov2023direct,shao2024deepseekmath}, improving formatting and reward-guided alignment~\cite{kumar2025llm,tie2025survey}. Recent systems like GPT-4~\cite{achiam2023gpt}, Claude-3.5~\cite{claude3}, and DeepSeek-R1~\cite{guo2025deepseek} showcase these techniques.
Despite recent advances, post-trained models still struggle with generalization under distribution shifts and novel tasks. While SFT stabilizes outputs but often overfits, RL improves adaptability~\cite{chu2025sft}, and combining both shows promise~\cite{guo2025deepseek}, making post-training vital for robust, aligned LLMs.

\paragraph{Test-time scaling.} Scaling inference-time compute improves performance without retraining. Approaches include beam search~\cite{lowerre1976harpy,graves2012sequence}, best-of-N sampling~\cite{sun2024fast}, and MCTS~\cite{coulom2006efficient}, alongside prompting methods like CoT~\cite{wei2022chain} and ToT~\cite{yao2023tree}. 
Strategically allocated test-time compute has proven effective for enhancing reasoning in both unimodal and multimodal models. Building on this, we apply CoT reasoning to 3D spatial reasoning by fine-tuning models to generate step-by-step rationales, providing both accurate and interpretable results.

\section{\modelname{}} \label{sec:method}

\subsection{Overview} \label{sec:method_overview}

\begin{figure}[t]
    \centering
    \includegraphics[width=\linewidth]{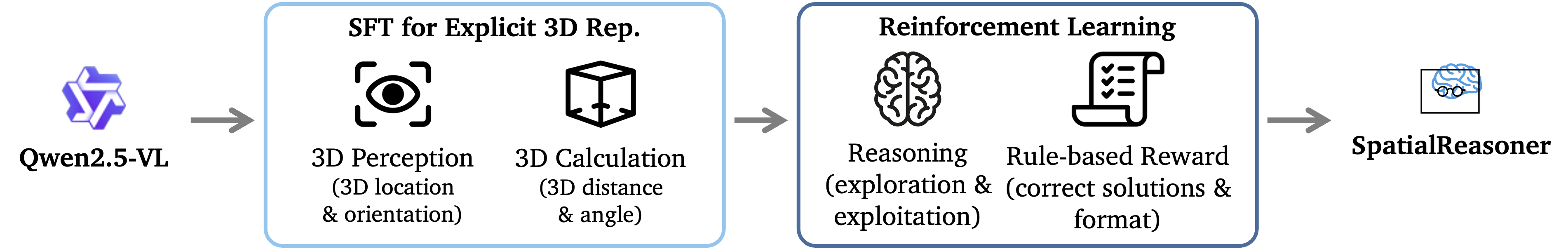}
    \caption{\textbf{Overview of our \modelname{} training.} We adopt a multi-stage training strategy: In Stage I, we apply supervised fine-tuning (SFT) to equip the LVLM with explicit 3D representations; then in Stage II, we leverage reinforcement learning (RL) to develop robust and generalizable 3D spatial reasoning built on explicit 3D representations.}
    \label{fig:training}
\end{figure}

In this section, we introduce our \modelname{} for explicit and generalizable 3D spatial reasoning. Our \modelname{} features two key designs: (1) explicit 3D representations that serves as interface to support multi-stage spatial reasoning, \textit{i.e.}, 3D parsing, computation, and reasoning (see Figure~\ref{fig:design}), and (2) generalizable spatial reasoning from multi-stage training (see Figure~\ref{fig:training}).

In Section~\ref{sec:explicit}, we present the explicit 3D representations and describe how our model is trained to predict and to interpret the 3D representations for spatial reasoning. Then in Section~\ref{sec:method_training} we discuss our training strategies, exploring standard supervised fine-tuning, reinforcement learning, as well as 3D-aware process rewards. Lastly we introduce our 3D-aware data generation pipeline and different variants of training data used at different stages in Section~\ref{sec:method_data}.

\begin{figure}[t]
    \centering
    \includegraphics[width=0.5\linewidth]{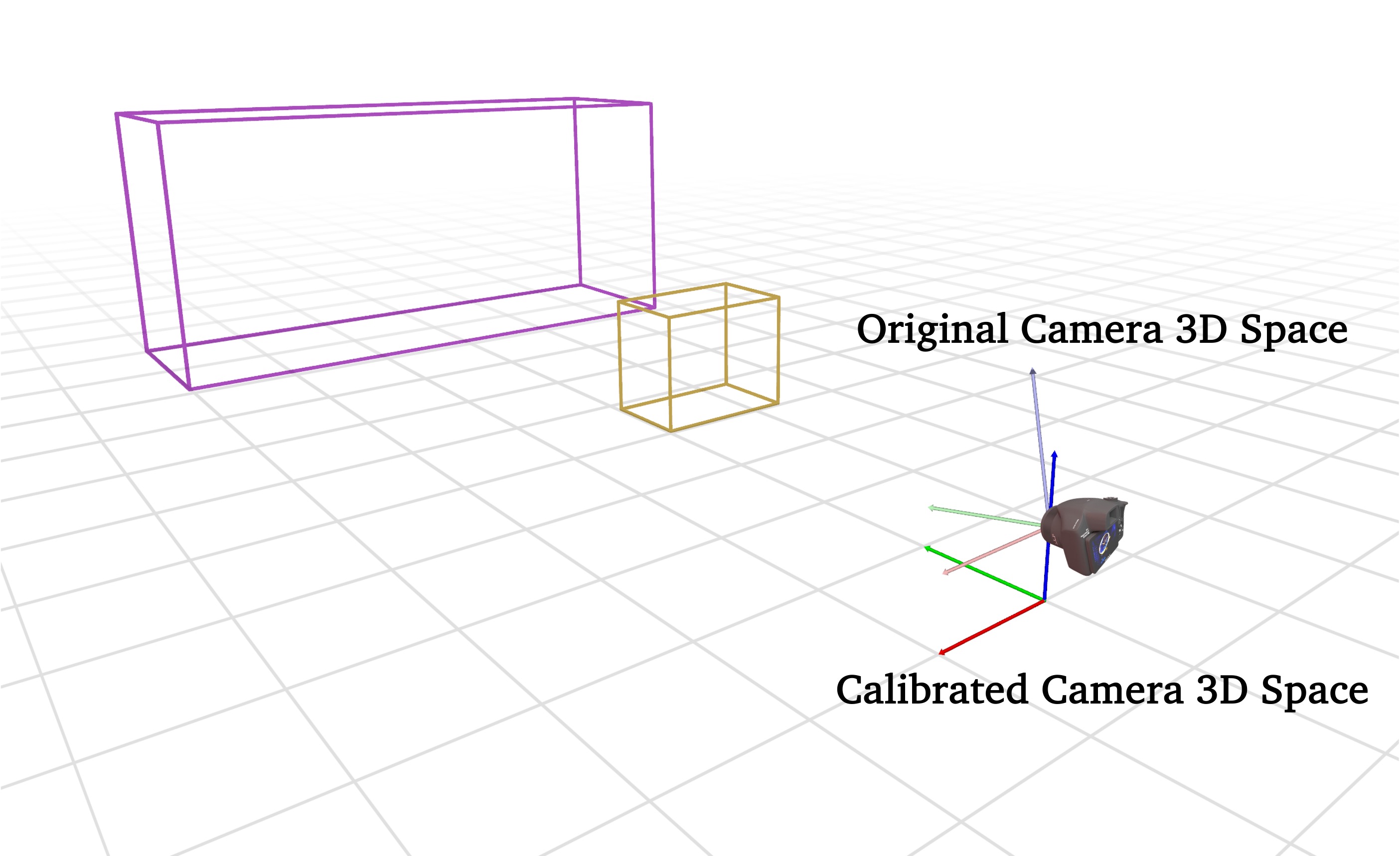}
    \caption{\textbf{Comparison between original and calibrated camera 3D space.} Our explicit 3D representations are defined within calibrated camera 3D space that simplifies subsequent 3D computations.}
    \label{fig:calibrated}
\end{figure}

\subsection{Learning Explicit 3D Representations} \label{sec:explicit}

Despite the improved spatial reasoning abilities achieved by 3D-aware LVLMs, such as SpatialRGPT~\cite{cheng2024spatialrgpt} and SpatialLLM~\cite{ma2025spatialllm}, and advanced proprietary models like Gemini 2.0~\cite{google2024gemini2}, these methods lack explicit 3D representations and instead rely on natural language to perform 3D spatial reasoning. For example, Gemini 2.0 predicts 3D object poses with descriptions like ``facing towards the viewer and slightly to its right'' and estimates 3D distances with phrases such as ``far behind some other object''. These natural language descriptions are inefficient and often not accurate enough for complex 3D spatial reasoning.

Therefore, we propose integrating LVLMs with explicit 3D representations, such as 3D locations and poses, to serve as an accurate and reliable interface shared across stages of 3D spatial reasoning (see Figure~\ref{fig:design}). Our \modelname{} can predict explicit 3D representations as intermediate results or take them as inputs to perform basic 3D computations or complex spatial reasoning tasks.

\paragraph{Explicit 3D representations.} We define explicit 3D representations in a calibrated camera 3D space, which is a standard camera 3D space calibrated with the extrinsics of the camera.
As illustrated in Figure~\ref{fig:calibrated}, the calibrated camera 3D space has its $z$-axis aligned with the $z$-axis of the 3D world space, and the origin on the $z$-axis is positioned close to the ground plane.
Although estimating 3D object locations is easier in original camera 3D space, adopting explicit 3D representations in calibrated camera 3D space offers many advantages for subsequent spatial reasoning: (1) the $z$-coordinates directly correspond to object heights, (2) estimating 3D spatial relationships such as ``above'' and ``below'' is largely simplified, and (3) objects often are on a plane parallel to the ground plane, which reduces many 3D spatial relationships to simpler 2D problems.

\paragraph{A unified interface for explicit 3D spatial reasoning.} Explicit 3D representations serve as an interface enabling coherent and accurate 3D spatial reasoning across stages (see Figure~\ref{fig:design}).
For 3D perception, the model predicts object locations and orientations as 3D vectors, then estimates explicit distances or angles based on these predictions. Finally, the model aggregates explicit 3D information from earlier stages to reason about and answer 3D spatial questions.

\subsection{Training Strategies} \label{sec:method_training}

\paragraph{Supervised fine-tuning (SFT).}
We adopt a two-stage post-training strategy to equip the model with explicit, generalizable 3D spatial reasoning capabilities.
In the first stage, SFT serves as a critical initialization step, aligning the pre-trained LVLM with curated 3D-annotated datasets.
By optimizing a maximum likelihood objective over paired input-output sequences, the model learns to identify objects in 3D scenes and predict explicit 3D representations such as object locations and orientations.
This structured supervision enables the model to expose interpretable intermediate reasoning traces during spatial computations—\eg/, calculating relative distances and angles.
However, due to its reliance on static demonstrations, SFT shows limited capacity to generalize beyond observed reasoning patterns.
As shown in Section~\ref{sec:analyses}, SFT-trained models tend to memorize spatial templates in the training set and struggle with novel compositions or combinatorial variations in 3D queries.

\paragraph{Reinforcement learning (RL).} To overcome this limitation, we further post-train the SFT-initialized model using reinforcement learning (RL) with rule-based rewards. 
Treating the spatial reasoning task as a sequential decision process, each reasoning step is framed as an action within a Markov Decision Process (MDP), and policy gradients are optimized using GRPO~\cite{shao2024deepseekmath}.
Importantly, we integrate a reward scheme that provides structured reward signals reflecting both the correctness of final answers and the coherence of intermediate 3D computation steps. 
This enables multi-turn optimization where the model learns to revise inaccurate inferences and explore alternative reasoning paths. 
Empirically, this RL stage substantially improves generalization to out-of-distribution 3D spatial questions (Table~\ref{tab:generalizability}), particularly in settings involving multi-object arrangements. 
The combined SFT+RL strategy therefore balances response formatting and perceptual grounding with adaptive reasoning and robustness, making it especially suited for complex 3D spatial understanding.

\paragraph{Reward and policy optimization design.}
To train \modelname{} via reinforcement learning, we design a composite reward scheme capturing both answer correctness and reasoning quality.
For outcome rewards, we use an accuracy reward—aligned with the multiple-choice evaluation metric from MMBench~\cite{liu2024mmbench}—which gives positive feedback only when the model selects the correct answer, and a format reward~\cite{guo2025deepseek} that encourages structured, readable outputs.
We also explore process rewards to assess whether models can learn structured reasoning without SFT. These include a reasoning steps reward, which promotes use of structured indicators (\eg/, ``First'', ``Next''), and a 3D-aware process reward that checks for necessary spatial terms like distance or orientation. The reward signal is then calculated through the accuracy of the presence of each term. Though not used in the final model, these rewards serve as diagnostics for emergent reasoning behavior.
Finally, we ablate the KL divergence term in GRPO and find that it harms training accuracy despite stabilizing output lengths (Section~\ref{sec:analyses}). We thus remove it in our final setting.

\subsection{Training Data} \label{sec:method_data}

To enable LVLMs to predict and reason with explicit 3D representations and to post-train LVLMs to solve various challenging 3D spatial reasoning questions, we generate a series of 3D-aware training data.
We extend the data generation pipeline in~\cite{chen2024spatialvlm,cheng2024spatialrgpt,ma2025spatialllm}. Our process begins with generating 3D pseudo-annotations, followed by optional human verification, and ends with constructing various VQAs and chain-of-thought reasoning based on the 3D pseudo-annotations.

\paragraph{Pseudo 3D ground-truths.} We extend the 3D pseudo-annotation pipeline proposed in~\cite{chen2024spatialvlm,ma2025spatialllm} and generate various 3D annotations, such as object category, 3D location, and 3D pose, on unlabeled images from the OpenImages dataset~\cite{OpenImages}. Based on the object-level 3D annotations, we then apply rule-based methods to derive ground-truth labels for a range of 3D spatial relationships.
Despite significant progress in visual foundation models for segmentation~\cite{ravi2024sam}, metric depth estimation~\cite{yang2024depth}, and object pose estimation~\cite{ma2024imagenet3d}, we notice many factual errors in the generated 3D ground-truth, particularly when mistakes (\textit{e.g.}, missing objects or inaccurate object poses) propagate through the data pipeline.
Therefore, we adopt a series of aggressive filtering steps to ensure the quality of our training data, including: (1) removing images with densely cluttered scenes, (2) excluding object categories that are difficult to segment or estimate pose for, and (3) discarding boundary cases that could lead to ambiguity.

\paragraph{Human verification.} Despite leveraging state-of-the-art visual foundation models in our data generation pipeline and applying multiple filtering strategies, the 3D pseudo-annotations remain susceptible to factual errors. Specifically, small issues such as missing objects or inaccurate 3D pose predictions propagate through later stages of the pipeline, leading to factual errors in the final spatial relationship pseudo-annotations. To assess the impact of data quality, we create a smaller but higher-quality dataset by manually verifying the correctness of the 3D pseudo-annotations.

\paragraph{Training data variants.} Based on the obtained 3D ground-truth, we can generate different variants of data for fine-tuning. Specifically, we consider the following:
(1) \textit{Basic3D-QA} consists of basic 3D perception and 3D computation question-answering data. This can be used to learn explicit 3D representations without training on various 3D spatial relationships considered in downstream tasks.
(2) \textit{SR-QA} contains visual question-answering pairs about various 3D spatial relationships, following previous 3D-aware datasets~\cite{cheng2024spatialrgpt,ma2025spatialllm}.
(3) \textit{SR-CoT} extends SR-QA and comprises chain-of-thought reasoning with explicit 3D representations. Questions are answered in a step-by-step manner.

\begin{table}[t]
    \centering
    \footnotesize
    \begin{tabular}{l p{2cm} p{2cm} p{2cm} p{2cm} p{2cm}}
        \toprule
        Method & Mean & Height & Location & Orientation & Multi-Object \\
        \midrule
        \textbf{\textit{Open-Sourced Generalist}} \\
        LLaVA-v1.5-7B~\cite{liu2024improved} & 38.1 & 39.1 & 46.9 & 28.7 & 34.7 \\
        LLaVA-Next-8B~\cite{li2024llavanext-strong} & 48.4 & 50.6 & 59.9 & 36.1 & 43.4 \\
        Cambrian-1-8B~\cite{tong2024cambrian} & 42.2 & 23.2 & 53.9 & 35.9 & 41.9 \\
        Qwen2.5-VL-3B-Instruct~\cite{Qwen2.5-VL} & 43.9 & 45.2 & 56.8 & 35.7 & 35.7 \\
        Qwen2.5-VL-7B-Instruct~\cite{Qwen2.5-VL} & 48.4 & 44.1 & 62.7 & 40.6 & 40.5 \\
        Qwen2.5-VL-72B-Instruct~\cite{Qwen2.5-VL} & 54.9 & 53.3 & 71.0 & 43.1 & 46.6 \\
        \midrule
        \textbf{\textit{Open-Sourced Specialist}} \\
        SpaceLLaVA~\cite{spacellava} & 42.0 & 49.3 & 54.4 & 27.6 & 35.4 \\
        SpatialBot~\cite{cai2024spatialbot} & 41.0 & 40.4 & 54.4 & 31.9 & 33.5 \\
        SpatialLLM~\cite{ma2025spatialllm} & 44.8 & 45.8 & 61.6 & 30.0 & 36.7 \\
        SpatialRGPT~\cite{cheng2024spatialrgpt} & 32.7 & \textbf{55.9} & 39.0 & 27.8 & 20.0 \\
        \textcolor{gray}{SpatialRGPT} \textcolor{gray}{\textit{w/} depth} ~\cite{cheng2024spatialrgpt} & \textcolor{gray}{48.4} & \textcolor{gray}{55.9} & \textcolor{gray}{60.0} & \textcolor{gray}{34.2} & \textcolor{gray}{42.3} \\
        \midrule
        \textbf{\textit{Proprietary}} \\
        GPT-4o-mini & 39.7 & 44.3 & 52.4 & 21.0 & 36.5 \\
        GPT-4o & 44.2 & 53.2 & 59.6 & 21.6 & 39.0 \\
        Claude 3.5 V Sonnet & 48.2 & 53.5 & 63.1 & 31.4 & 41.3 \\
        Gemini 2.0 Flash & 49.8 & 49.7 & 68.9 & 32.2 & 41.5 \\
        Gemini 2.0 Flash (thinking) & 51.1 & \textbf{53.0} & 67.1 & 35.8 & 43.6 \\
        QwenVLMax & 52.0 & 45.1 & 70.7 & 37.7 & 44.8 \\	
        \midrule
        \textbf{\textit{Ours}} \\
        \cellcolor{gray!10}\modelname{}-Zero & \cellcolor{gray!10}54.0 & \cellcolor{gray!10}46.4 & \cellcolor{gray!10}67.3 & \cellcolor{gray!10}48.4 & \cellcolor{gray!10}47.2\\ 
        \cellcolor{gray!10}\modelname{}-SFT & \cellcolor{gray!10}\underline{58.3} & \cellcolor{gray!10}51.9 & \cellcolor{gray!10}\underline{73.5} & \cellcolor{gray!10}\underline{50.7} & \cellcolor{gray!10}\underline{50.3}\\ 
        \cellcolor{gray!10}\modelname{} & \cellcolor{gray!10}\textbf{60.3} & \cellcolor{gray!10}\underline{52.5} & \cellcolor{gray!10}\textbf{75.2} & \cellcolor{gray!10}\textbf{55.2} & \cellcolor{gray!10}\textbf{51.8}\\
        \bottomrule
    \end{tabular}
    \caption{\textbf{Comparison with previous state-of-the-art methods on 3DSRBench~\cite{ma20243dsrbench}.} Our \modelname{} outperforms previous open-source and proprietary methods on challenging 3D spatial reasoning problems in 3DSRBench.}
    \label{tab:sota}
\end{table}

\section{Results} \label{sec:experiments}

\subsection{Experimental Setup}

\paragraph{Baselines.} We compare our \modelname{} with the following three types of baseline models.
(1) \textit{Open-sourced generalist models}: such as LLaVA~\cite{liu2024improved}, Cambrian-1~\cite{tong2024cambrian}, and Qwen2.5~\cite{Qwen2.5-VL} that are trained on general vision-language data.
(2) \textit{Open-sourced specialist models}: We evaluate SpaceLLaVA~\cite{spacellava}, which is a public re-implementation of SpatialVLM~\cite{chen2024spatialvlm}, SpatialBot~\cite{cai2024spatialbot} that enhances fine-grained spatial reasoning and robot control by utilizing both RGB and depth images, and SpatialLLM~\cite{ma2025spatialllm} that fine-tunes a LLaVA model with multi-stage 3D-informed training. Note for fair comparison, we evaluate SpatialBot with RGB inputs only.
(3) \textit{Proprietary models} such as GPT-4o~\cite{achiam2023gpt} and Claude 3.5~\cite{claude3.5} that have been trained on abundant web-scale data and for Gemini 2.0~\cite{google2024gemini2}, additional 3D-aware post-training.


\paragraph{Evaluation benchmarks.} We evaluate spatial reasoning abilities of various models on three spatial reasoning benchmarks. \textit{3DSRBench}~\cite{ma20243dsrbench} is a comprehensive 3D spatial reasoning benchmark with 2,100 questions and studies various 3D awareness and reasoning abilities with a robust evaluation setup.
\textit{CVBench}~\cite{tong2024cambrian} is a vision-centric benchmark that assesses models at classic vision tasks with a range of 2D and 3D understanding VQAs. In this work, we focus exclusively on 3D-related questions, \textit{i.e.}, CVBench-3D, as 2D left-right relationships can lead to ambiguity with 3D left-right questions considered in 3DSRBench.
\textit{GQA}~\cite{hudson2019gqa} is a widely adopted benchmark that studies visual reasoning and compositional question answering on a range of spatial relationships between objects.

\begin{table}[t]
    \centering
    \footnotesize
    \begin{tabular}{l p{1.1cm} p{1.1cm} | p{1.1cm} p{1.1cm} p{1.1cm} p{1.1cm} p{1.1cm} p{1.1cm}}
        \toprule
        & \multicolumn{2}{c|}{CV-Bench-3D} & \multicolumn{6}{c}{GQA} \\
        Method & Depth & Distance & Mean & Choose & Compare & Logical & Query & Verify \\
        \midrule
        Qwen2.5-VL-7B-Instruct~\cite{Qwen2.5-VL} & 82.5 & \textbf{83.2} & 58.8 & 82.7 & 71.5 & 78.9 & 40.6 & 82.5\\
        %
        \cellcolor{gray!10}\modelname{}-Zero & \cellcolor{gray!10}77.5 & \cellcolor{gray!10}\underline{81.8} & \cellcolor{gray!10}60.2 & \cellcolor{gray!10}81.6 & \cellcolor{gray!10}67.9 & \cellcolor{gray!10}\underline{79.2} & \cellcolor{gray!10}43.8 & \cellcolor{gray!10}82.1\\
        \cellcolor{gray!10}\modelname{}-SFT & \cellcolor{gray!10}\underline{85.2} & \cellcolor{gray!10}71.5 & \cellcolor{gray!10}\textbf{62.0} & \cellcolor{gray!10}\underline{82.8} & \cellcolor{gray!10}\underline{72.2} & \cellcolor{gray!10}\textbf{81.4} & \cellcolor{gray!10}\textbf{45.5} & \cellcolor{gray!10}\textbf{83.2}\\
        \cellcolor{gray!10}\modelname{} & \cellcolor{gray!10}\textbf{87.3} & \cellcolor{gray!10}73.3 & \cellcolor{gray!10}\underline{61.8} & \cellcolor{gray!10}\textbf{83.2} & \cellcolor{gray!10}\textbf{81.1} & \cellcolor{gray!10}71.8 & \cellcolor{gray!10}\underline{45.2} & \cellcolor{gray!10}\underline{82.9}\\
        \bottomrule
    \end{tabular}
    \caption{\textbf{Performance on CV-Bench-3D and GQA.} Our \modelname{} also improves the spatial reasoning performance on GQA~\cite{hudson2019gqa} and depth-related questions in CVBench-3D~\cite{tong2024cambrian}. For distance-related questions, unlike Qwen2.5 that exhibits excessive dependence on 2D shortcuts, our \modelname{} employs rigid 3D spatial reasoning and achieves compelling performance. Regarding the performance on distance-related questions, see Section~\ref{sec:advancing} and Section~\ref{sec:supp_spurious} for detailed discussions.}
    \label{tab:other_bench}
\end{table}

\begin{table}[t]
    \centering
    \footnotesize
    \begin{tabular}{l p{2cm} p{2cm} p{2cm} p{2cm} | p{2cm}}
        \toprule
        & & \multicolumn{3}{c|}{In-Distribution} & \multicolumn{1}{c}{Novel} \\
        Method & Mean & Height & Location & Orientation & Multi-Object \\
        \midrule
        Qwen2.5-VL-7B-Instruct~\cite{Qwen2.5-VL} & 48.4 & 44.1 & 62.7 & 40.6 & 40.5\\
        \cellcolor{gray!10}\modelname{}-Zero & \cellcolor{gray!10}\underline{53.7} & \cellcolor{gray!10}40.6 & \cellcolor{gray!10}68.4 & \cellcolor{gray!10}\underline{50.2} & \cellcolor{gray!10}\textbf{46.6}\\
        \cellcolor{gray!10}\modelname{}-SFT & \cellcolor{gray!10}52.2 & \cellcolor{gray!10}\underline{44.9} & \cellcolor{gray!10}\underline{69.5} & \cellcolor{gray!10}48.9 & \cellcolor{gray!10}40.0\\
        \cellcolor{gray!10}\modelname{} & \cellcolor{gray!10}\textbf{56.4} & \cellcolor{gray!10}\textbf{52.5} & \cellcolor{gray!10}\textbf{72.6} & \cellcolor{gray!10}\textbf{54.1} & \cellcolor{gray!10}\underline{43.4}\\
        \bottomrule
    \end{tabular}
    \caption{\textbf{
    Evaluation of generalization ability by finetuning on simpler (in-distribution) 3D spatial reasoning questions and evaluating on complex (novel) questions types unseen during training.} Our \modelname{}-Zero and \modelname{} demonstrates superior zero-shot generalization, indicating RL fosters more robust and flexible reasoning than SFT.}
    \label{tab:generalizability}
\end{table}

\subsection{Advancing 3D Spatial Reasoning}\label{sec:advancing}
We evaluate \modelname{}~across 3DSRBench~\cite{ma20243dsrbench}, CVBench-3D~\cite{tong2024cambrian}, and GQA~\cite{hudson2019gqa} to systematically assess 3D perception, computation, and reasoning capabilities. 
As shown in Table~\ref{tab:sota}, \modelname{}~achieves a new state-of-the-art mean accuracy of 60.3\% on 3DSRBench, substantially outperforming prior open-source and proprietary models. 
Compared to Gemini 2.0 Flash (49.8\%) and Claude 3.5 Sonnet (48.2\%), our model demonstrates significant gains across all dimensions, with particularly strong improvements in Location (+6.3\%) and Orientation (+14.6\%) questions relative to the second-best model, reflecting better 3D perception and computation abilities. 
Furthermore, \modelname{}~attains 51.8\% (+8.2\%) accuracy on Multi-Object questions, highlighting its ability to reason about complex spatial interactions involving multiple entities.

On CVBench-3D and GQA (see Table~\ref{tab:other_bench}), \modelname{}~maintains strong generalization, excelling particularly in depth-related spatial perception with an 87.3\% accuracy.
In GQA, our model shows improved performance in the Compare category, further supporting enhanced multi-object reasoning capabilities. 
These gains validate the strength of our multi-stage training pipeline: SFT grounds the model with explicit 3D representations for accurate perception and basic spatial computations, while RL promotes adaptive multi-step reasoning and generalization to novel 3D spatial configurations. Collectively, these advances establish \modelname{}~as a new benchmark for vision-language models in 3D spatial reasoning tasks.

Despite this, we observe a noticeable drop in the performance of \modelname{} on CVBench-3D distance questions. This stands in contrast to the significant improvement in performance on similar questions in 3DSRBench (``multi-object-closer-to''), where performance increases from 34.3\% to 70.9\%. We attribute this gap to the abundant shortcuts present in distance-related questions in CVBench-3D, where LVLMs tend to exploit 2D spurious correlations as a shortcut for 3D spatial reasoning. In contrast, our \modelname{} builds on explicit 3D representations, enabling it to better tackle challenging 3D spatial reasoning questions. Please refer to Section~\ref{sec:supp_spurious} for detailed discussions.

\subsection{Analyses and Findings} \label{sec:analyses}

\begin{table}[t]
    \centering
    \footnotesize
    \begin{minipage}{0.45\textwidth}
        \centering
        \begin{tabular}{p{5.7cm} p{1.00cm}}
            \toprule
            Method & Mean \\
            \midrule
            Qwen2.5-VL-7B-Instruct~\cite{Qwen2.5-VL} & 48.4 \\
            \cellcolor{gray!10}\modelname{}-SFT & \cellcolor{gray!10}58.3 \\
            \cellcolor{gray!10}\modelname{}-SFT (\textit{w/o} explicit 3D rep.) & \cellcolor{gray!10}51.9\\
            \bottomrule
        \end{tabular}
        \caption{\textbf{Comparisons between \modelname{} with and without explicit 3D representations and CoT reasoning.} Results highlight the benefits of explicit 3D representations as an interface to support enhanced 3D spatial reasoning.}
        \label{tab:ablation-a}
    \end{minipage}
    \hfill
    \begin{minipage}{0.5\textwidth}
        \centering
        \begin{tabular}{p{6.4cm} p{1.00cm}}
            \toprule
            Method & Mean \\
            \midrule
            Qwen2.5-VL-7B-Instruct~\cite{Qwen2.5-VL} & 48.4 \\
            \cellcolor{gray!10}\modelname{}-SFT & \cellcolor{gray!10}58.3 \\
            \cellcolor{gray!10}\modelname{}-SFT (+HQ SFT) & \cellcolor{gray!10}54.7 \\
            \midrule
            \cellcolor{gray!10}\modelname{}-Zero & \cellcolor{gray!10}54.0 \\
            \cellcolor{gray!10}\modelname{}-Zero (\textit{w/} KL) & \cellcolor{gray!10}52.4 \\
            \cellcolor{gray!10}\modelname{}-Zero (\textit{w/} 3D Rwd) & \cellcolor{gray!10}54.6\\
            \bottomrule
        \end{tabular}
        \caption{\textbf{Ablation study on various design choices in RL and SFT.} Notably with 3D-aware rewards, \modelname-Zero produces coherent chain-of-thought reasoning on explicit 3D representations and improves benchmark performance.}
        \label{tab:ablation-b}
    \end{minipage}
\end{table}

\paragraph{Generalization abilities.} 
Recent studies have shown that while SFT helps stabilize model outputs, it tends to cause memorization, limiting generalization, especially when encountering out-of-distribution (OOD) variations~\cite{chu2025sft}. In contrast, RL, particularly when using outcome-based rewards, enhances a model's ability to learn transferable principles, enabling better adaptation to unseen tasks and improving perceptual capabilities. Motivated by these findings, we design an experiment to assess how different post-training strategies—SFT, RL, and their combination—affect generalization in spatial reasoning, specifically on multi-object scenes, which are often more challenging due to increased visual and relational complexity.

To evaluate this, we removed all multi-object training data and trained three models: \modelname-Zero (RL-only), \modelname-SFT (SFT-only), and \modelname~(SFT+RL). As shown in Table~\ref{tab:generalizability}, the multi-object performance of \modelname-Zero (46.6\%) substantially surpasses that of \modelname-SFT (40.0\%), highlighting that RL training alone provides better zero-shot generalization to multi-object reasoning tasks than SFT alone. Furthermore, combining SFT and RL (\modelname, 43.4\%) further improves over SFT, but does not fully match the RL-only model's multi-object performance when trained without direct supervision. This trend echoes the hypothesis that SFT tends to overfit to the specific distributions seen during training, while RL promotes more flexible reasoning capabilities.

When multi-object training examples exist (Table~\ref{tab:sota}), SFT can simply memorize the structure and examples of multi-object scenes through direct supervision, resulting in reasonable performance by relying on pattern matching (50.3\% vs. 40.5\%). However, when multi-object examples are removed, SFT performance drops sharply (40.0\% vs. 50.3\%), while RL-tuned models continue to generalize well (46.6\% vs. 47.2\%). This suggests that RL is not merely memorizing the training-specific distributions but is instead internalizing transferable and compositional reasoning capabilities, enabling robust generalization to complex unseen multi-object scenarios.

Beyond the multi-object class performance, we also analyze the overall mean performance changes across settings. As shown in Table~\ref{tab:sota} and Table~\ref{tab:generalizability}, \modelname-Zero maintains almost the same mean performance (54.0\% vs. 53.7\%), indicating that RL training enables strong generalization even when multi-object examples are removed. In contrast, \modelname-SFT suffers a substantial mean performance degradation (58.3\% vs. 52.2\%), reflecting its reliance on memorizing specific training distributions. Overall, these results reinforce the conclusion that outcome-based RL training enables models to acquire more robust and transferable reasoning capabilities, while SFT alone remains vulnerable to overfitting to seen distributions.

\insightbox{While SFT helps stabilize outputs, it tends to overfit training distributions, whereas outcome-based RL encourages the development of transferable reasoning strategies that enable robust generalization to unseen scenarios.}

\begin{figure}[t]
    \centering
    \begin{minipage}{0.48\textwidth}
        \centering
        \includegraphics[width=\textwidth]{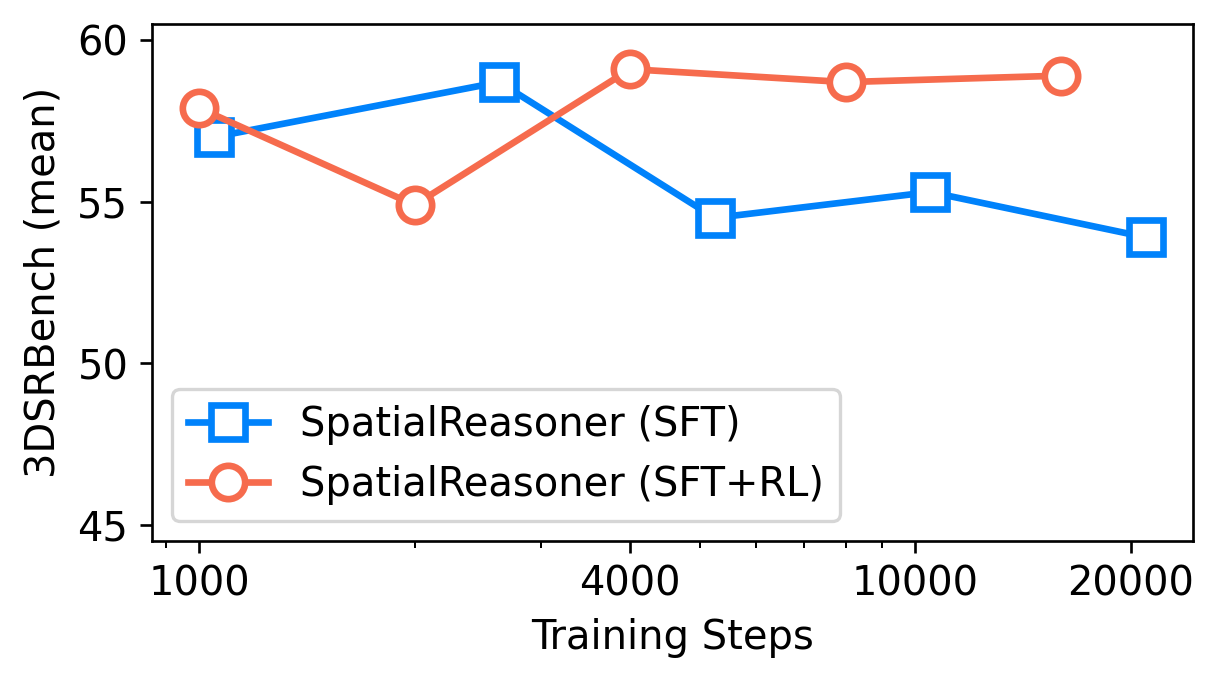}
        \caption{\textbf{Scaling of training computation.} \modelname{}-SFT may overfit to training data while \modelname{} trained with RL retains a stable and competitive performance.}
        \label{fig:scaling_computation}
    \end{minipage}
    \hfill
    \begin{minipage}{0.48\textwidth}
        \centering
        \includegraphics[width=\textwidth]{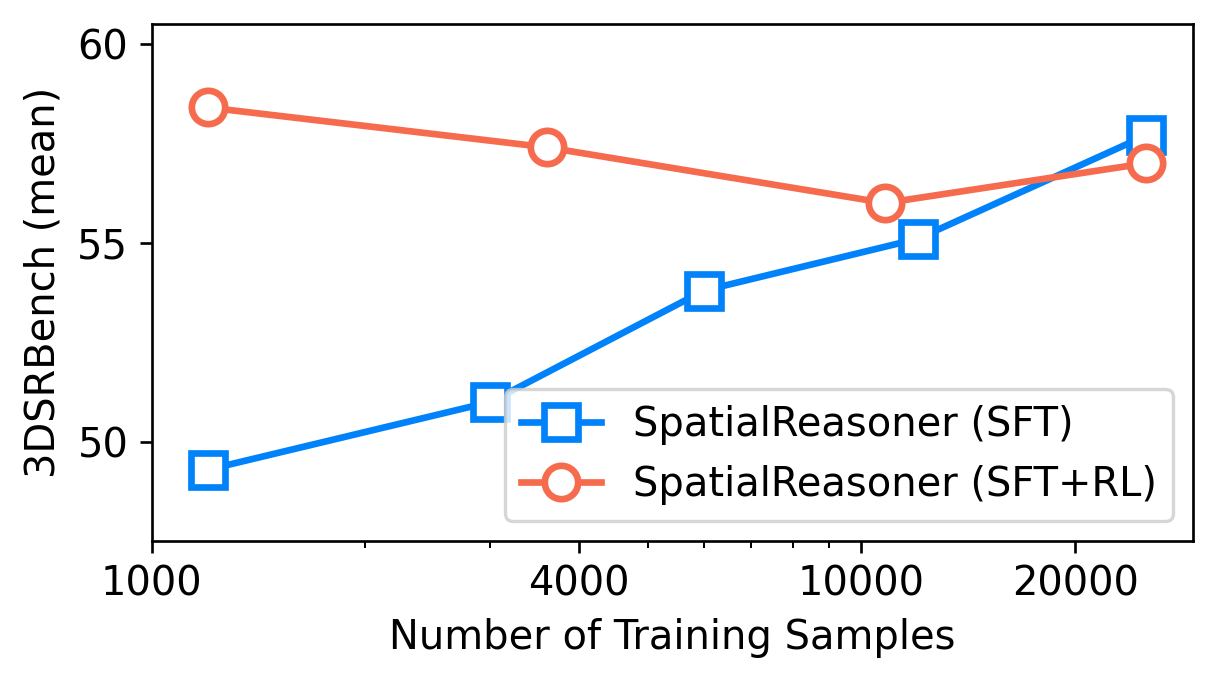}
        \caption{\textbf{Scaling of training data by mixing 1.2K human-verified with 24K unverified data.} Results show SFT remains a compelling choice when scaling with potentially noisy training data.}
        \label{fig:scaling_data}
    \end{minipage}
\end{figure}

\paragraph{Scaling of training computation.} We ablate on the scaling of training computation for \modelname{}-SFT and \modelname~in Figure~\ref{fig:scaling_computation}. Results show that with more training steps, \modelname{}-SFT starts to overfit and exhibit decreased performance, while \modelname{} trained with RL retains a stable and competitive performance.

\paragraph{Scaling of training data.} A major challenge for 3D-related tasks is the availability of high-quality data with 3D (pseudo-)annotations.
Since annotating 3D data is often time-consuming and requires certain expertise, real-world images with accurate 3D annotations are often limited in quantity.
On the other hand, synthetic data with 3D ground-truths or real-world images with automated pseudo-annotations are more abundant in quantity, but they inevitably introduce challenges such as domain gaps and noisy supervision.

\begin{figure}[t]
    \centering
    \includegraphics[width=\linewidth]{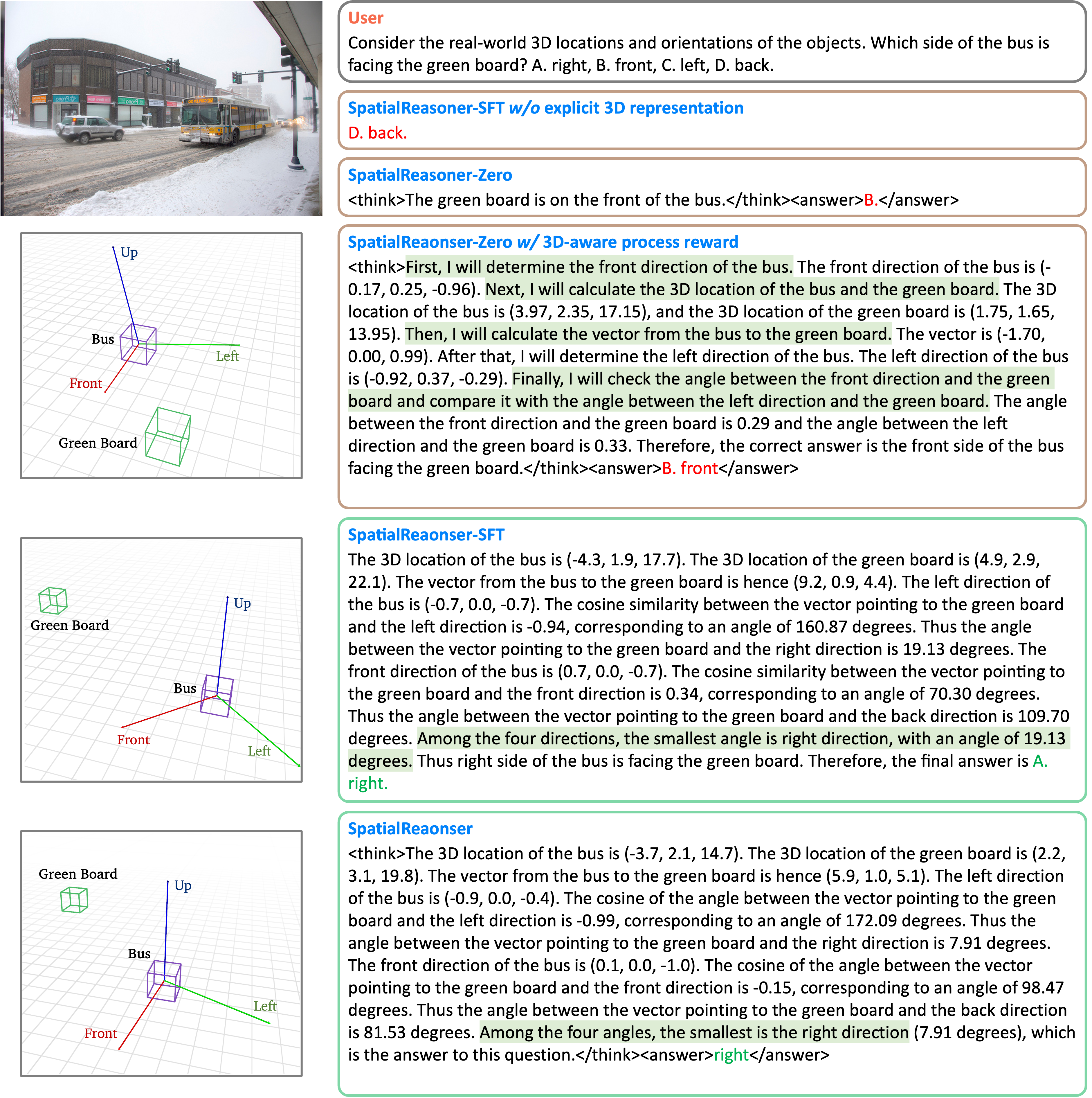}
    \caption{\textbf{Qualitative comparisons.} Our explicit 3D spatial reasoning improves interpretability over the baseline that relies on shallow shortcuts.}
    \label{fig:qualitative}
\end{figure}

To study the data scaling properties of our \modelname{} with or without reinforcement learning, we experiment on a range of training data sizes by mixing 1.2K human verified 3D data with 24K unverified 3D data. We visualize the data scaling results in Figure~\ref{fig:scaling_data} and find that \modelname{} trained with RL achieves the best performance when only 1.2K human verified data is used, which is in contrast to \modelname{}-SFT that scales effectively with the availability of more unverified 3D data.
This shows that SFT is more robust to noisy pseudo-annotations and benefits from data scaling, while fewer but higher-quality data are desirable for RL.

\insightbox{When developing 3D-aware LVLMs, SFT offers a more scalable approach given the availability of abundant, albeit possibly noisy, 3D pseudo-annotations, whereas RL benefits more from high-quality 3D data that are often limited in quantity.}

\paragraph{Ablation study on process reward modeling.} 
We conduct an ablation study to assess whether explicit 3D reasoning trajectory can emerge purely from process reward modeling, without relying on costly curated chain-of-thought supervision. 
Through qualitative analysis (Figure~\ref{fig:qualitative}), we observe that \modelname{}-Zero trained with 3D-aware process rewards can independently generate structured multi-step reasoning, correctly grounding its inferences through 3D locations, orientation estimation, and relational comparisons—despite not being exposed to manually crafted reasoning examples. 
Compared to \modelname{}-Zero, which often produces shortcut answers without sufficient process grounding, the enhanced model demonstrates a clear improvement in interpretability and reasoning fluency.
Quantitatively, as shown in Table~\ref{tab:ablation-b}, incorporating 3D-aware process rewards leads to a performance increase from 54.0\% to 54.6\%, highlighting that outcome-driven reward shaping can foster coherent reasoning without relying on costly human-curated chains of thought.
For the final \modelname{}~model, however, we rely on SFT to establish desirable reasoning processes, given its superior performance compared to process reward modeling alone.

\insightbox{Without abundant high-quality CoT reasoning data for SFT, our process reward modeling can effectively induce interpretable multi-step reasoning trajectories and enhance model performance.}

\paragraph{Ablation study on explicit 3D representations.} To study the importance of explicit 3D representations and step-by-step 3D computations, we train a variant of \modelname{} on the same data but without explicit 3D representations. Results in Table~\ref{tab:ablation-a} show that explicit 3D representations can enhance 3D spatial reasoning by a wide margin.

\begin{figure}[t]
    \centering
    \begin{subfigure}{0.48\textwidth}
    \includegraphics[width=\linewidth]{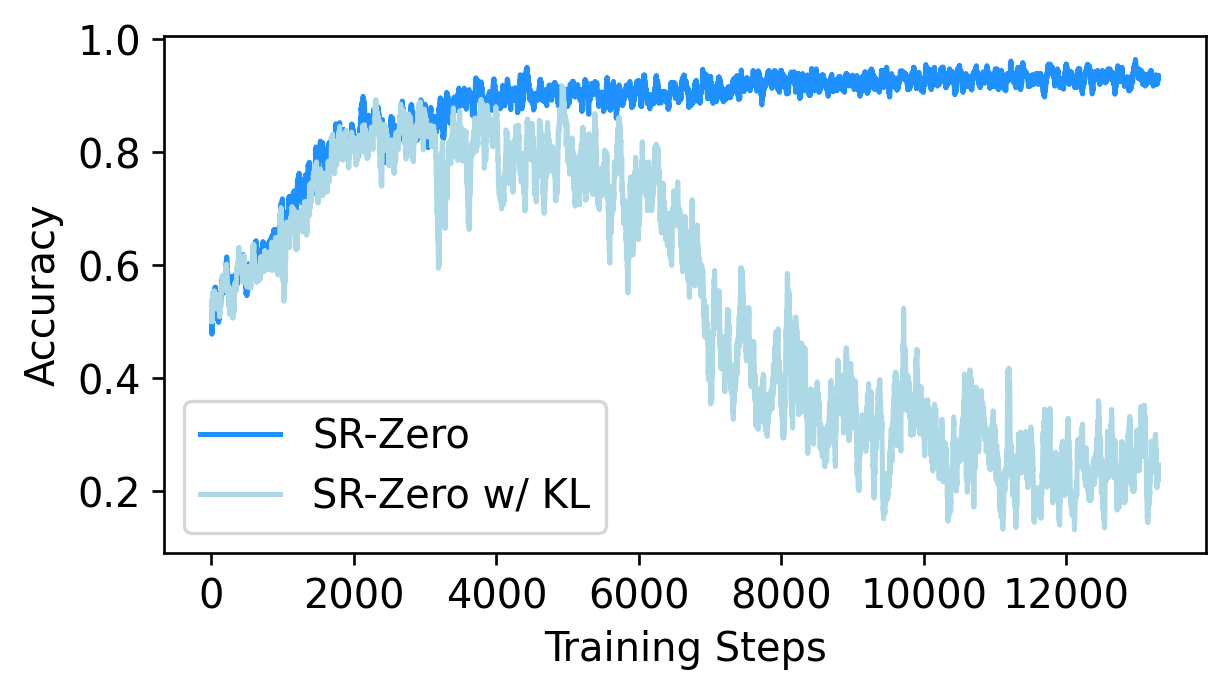}
    \caption{Accuracy of \modelname{}-Zero \textit{w/} and \textit{w/o} KL divergence.}
    \label{fig:curve_accuracy}
    \end{subfigure}
    \hfill
    \begin{subfigure}{0.48\textwidth}
    \includegraphics[width=\linewidth]{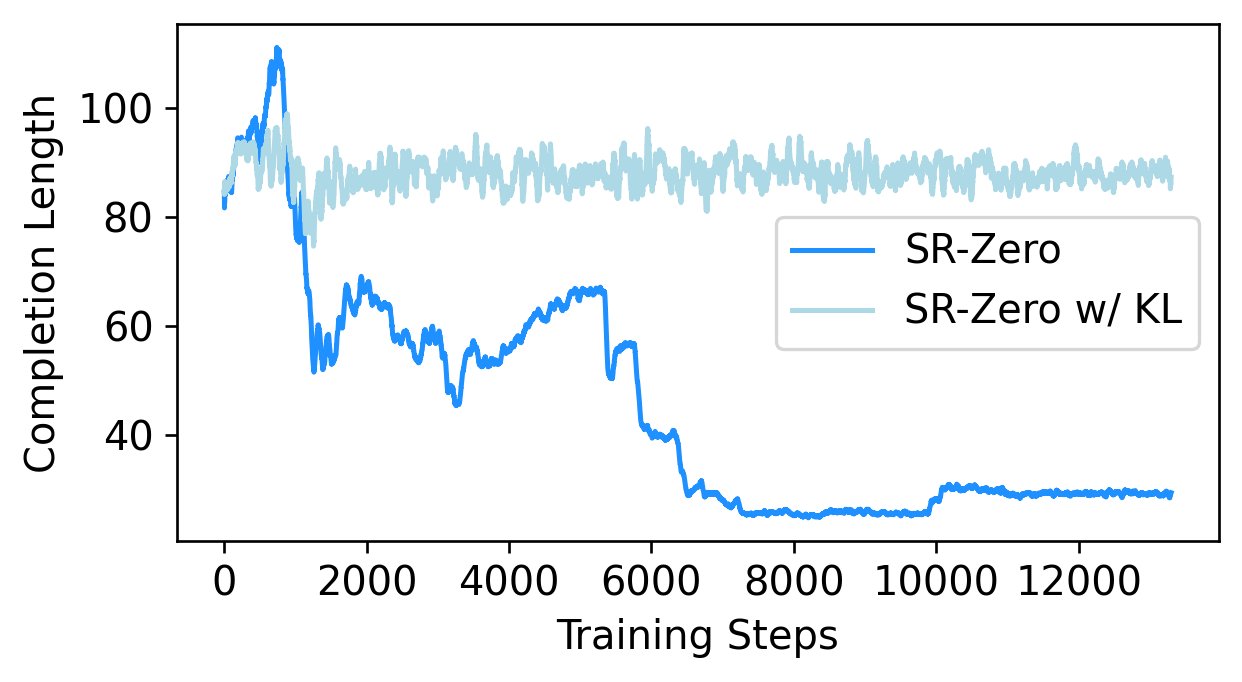}
    \caption{Completion length of \modelname{}-Zero \textit{w/} and \textit{w/o} KL divergence.}
    \label{fig:curve_length}
    \end{subfigure}
    \caption{\textbf{Training curve between \modelname{}-Zero and \modelname{}-Zero \textit{w/} KL.}}
    \label{fig:curve}
\end{figure}

\paragraph{Ablation study on KL divergence.}
We investigate the role of KL divergence regularization in GRPO and find that, although it traditionally serves to constrain policy drift and maintain linguistic stability during reward optimization, it significantly impairs learning in our setting.
As shown in Figure~\ref{fig:curve}, adding a KL constraint leads to an eventual collapse in training accuracy, while the model trained without KL divergence achieves stable improvements, reaching nearly perfect accuracy.
Compared to the stable completion length under KL regularization, the reduced completion length without KL might arise from mitigated length bias~\cite{liu2025understanding}, indicating improved token efficiency.
Quantitatively, as summarized in Table~\ref{tab:ablation-b}, \modelname{}-Zero achieves a higher accuracy of 54.0\% compared to 52.4\% when KL divergence is included. Based on these observations, we remove the KL divergence term from GRPO to promote better alignment with correct spatial reasoning answers.

\paragraph{Ablation Study on SFT+RL vs. SFT+SFT}
To isolate the effect of RL and verify that performance gains are not merely from additional data exposure, we compare sequential SFT+RL against SFT+SFT. As shown in Table~\ref{tab:ablation-b}, while SpatialReasoner-SFT achieves 58.3\% accuracy, applying a second round of SFT (SFT+SFT) degrades performance to 54.7\%. In contrast, switching to RL after SFT (SFT+RL) improves accuracy to 60.3\%. These results indicate that repeated SFT exacerbates overfitting to training patterns, whereas RL effectively builds on the structured outputs from SFT to promote more adaptive and generalizable 3D spatial reasoning.

\begin{table}[t]
    \centering
    \footnotesize
    \begin{tabular}{lccccccccc}
        \toprule
        & \multicolumn{2}{c}{3D Perception} & & \multicolumn{3}{c}{3D Reasoning} \\
        \cmidrule{2-3} \cmidrule{5-7}
        Method & Orientation ($\uparrow$) & Location ($\downarrow$) & & Angle ($\uparrow$) & Distance ($\downarrow$) & Depth ($\downarrow$) \\
        \midrule
        \modelname-SFT & 35.5 & 0.91 & & 55.0 & 0.17 & 0.13 \\
        \modelname & 31.0 & 1.05 & & 52.5 & 0.19 & 0.25 \\
        \bottomrule
    \end{tabular}
    \caption{\textbf{Studying failure modes of \modelname.} We observe that 3D reasoning can estimate angles, distances, and depths a lot more accurate than 3D perception that predicts orientations and locations from visual features.}
    \label{tab:failure}
\end{table}

\subsection{Interpreting Failure Modes}

Explicit 3D representations not only enhance 3D spatial reasoning, but also improve the interpretability of vision language systems~\cite{wang20233d}. We broadly categorize the 3D spatial reasoning into two stages: 3D perception, which parses crucial 3D information from the image input, and 3D reasoning, which computes key 3D metrics and derives the final answer.

To quantitatively study how models behave in the two stages, we manually evaluate the 3D pseudo-annotations generated by our data pipeline and obtain a total of 300 questions with human verified 3D answers. For 3D perception, we consider two types of questions that consider the 3D location and 3D orientation of the object. For 3D reasoning, models need to estimate distances, depths, and angles given 3D information such as locations and directions. Specifically, angles and directions are evaluated with prediction accuracy with a threshold of $\pi/6$ (30 degrees), while locations and distances are evaluated with mean error.

From the results in Table~\ref{tab:failure}, we observe that 3D reasoning can estimate angles, distances, and depths a lot more accurate than 3D perception that predicts orientations and locations from visual features. This observation aligns with findings from the qualitative examples where we visualize the predictions in a simulation system. We conclude that the majority of the errors made in downstream VQA tasks still stem from errors in 3D perception.
Moreover, RL slightly hurts the accuracies of 3D perception and reasoning. This is because the RL rewards only consider the format of the reasoning or the correctness of the final answer, ignoring all intermediate explicit 3D representations.

\section{Conclusions} \label{sec:conclusions}
In this paper, we introduced \modelname{}, a novel large vision-language model (LVLM) that performs explicit and generalizable 3D spatial reasoning by predicting and leveraging intermediate 3D representations across perception, computation, and reasoning stages. 
Through a two-stage post-training pipeline combining supervised fine-tuning and reinforcement learning, our method significantly advances the spatial reasoning abilities of LVLMs, achieving new state-of-the-art results across multiple benchmarks while maintaining strong generalization to novel 3D reasoning tasks.
By systematically analyzing model behaviors, we demonstrate that reasoning grounded in explicit 3D representations not only improves accuracy but also offers interpretable reasoning traces and highlights bottlenecks in 3D perception.
Our study underscores the importance of structured reasoning and adaptive training strategies for robust 3D understanding, opening new opportunities for future research at the intersection of visual perception, geometric computation, and multimodal reasoning.

\paragraph{Limitations.} Our system has a few limitations. First, despite the use of reinforcement learning (RL), supervised finetuning data with chain-of-thought reasoning and explicit 3D representations remains crucial as a warm-up. Training with RL alone--\textit{i.e.}, \modelname-Zero--fails to produce high-quality reasoning processes.
Second, \modelname{} consistently performs explicit multi-step 3D reasoning, regardless of the difficulty of the problems. Ideally the model should adopt a hybrid strategy--leveraging commen sense and visual cues for simpler questions while reserving explicit 3D computation and reasoning for more challenging problems. This would retain many strong and generalizable knowledge from the base Qwen2.5-VL model while improving inference efficiency.

\FloatBarrier

\bibliographystyle{abbrvnat}
\nobibliography*
\bibliography{document}

\clearpage

\appendix
\onecolumn
\section*{Appendix}

\section{Implementation detail}
We train \modelname{} using different combinations of curated datasets and training objectives. 
Starting from the Qwen2.5-VL-7B~\cite{Qwen2.5-VL} base model, we first apply SFT with 24k curated SR-CoT data alongside 24k randomly sampled LLaVA~\cite{liu2023visual} data, resulting in \modelname{}-SFT.
Next, we further train \modelname{}-SFT with RL using 1.2k SR-QA examples, leading to our final \modelname{}. For comparison, if we instead fine-tune \modelname{}-SFT with SFT on the same 1.2k SR-QA set, we obtain \modelname{}-SFT (+HQ SFT), serving as an ablation study baseline.
In parallel, we directly fine-tune the base model with RL using only the 1.2k SR-QA data without prior SFT, resulting in \modelname{}-Zero.
Additionally, to investigate whether explicit 3D perception capabilities can enable the model to self-organize reasoning trajectories under process rewards, we train the base model with SFT using 12k Basic3D-QA data alongside 12k randomly sampled LLaVA data before applying RL training on the 1.2k SR-QA data, yielding \modelname{}-Zero (\textit{w/} 3D Rwd).

We conduct all training experiments using 4×NVIDIA H100 80GB HBM3 GPUs. For SFT, we train the model for 10 epochs (approximately 20K steps with a batch size of 6) on the combined 24k SR-CoT and 24k LLaVA datasets. For RL training, we train for 100 epochs (approximately 13K steps with a batch size of 12) on the 1.2k SR-QA dataset, using 1 GPU with vLLM~\cite{kwon2023efficient} for efficient inference acceleration.
For the experiments withholding multi-object training examples, we double the number of training epochs to compensate for the reduced training set size.
We set the learning rate to 5e-6 for SFT and 5e-7 for RL, both following a cosine learning rate scheduler with a warm-up ratio of 0.1.
In the KL divergence ablation study, we set the KL penalty weight to 0.04.
We monitored the model every 1K training steps and reported results based on the best-performing checkpoint.

\section{2D Reasoning as a Shortcut} \label{sec:supp_spurious}

As shown in Table~\ref{tab:sota} and Table~\ref{tab:other_bench}, our \modelname{} outperforms previous open-source and proprietary models on 3DSRBench~\cite{ma20243dsrbench}, and achieves notable improvements on GQA~\cite{hudson2019gqa} (from 58.8\% to 61.8\%) and depth-related questions in CVBench-3D~\cite{tong2024cambrian} (from 82.5\% to 87.3\%).
%
%
However, if we focus specifically on multi-object 3D distance-related questions in CVBench-3D~\cite{tong2024cambrian} and 3DSRBench~\cite{ma20243dsrbench}, we observe contradictory results: \modelname{} achieves a substantial improvement of 21.5\% on 3DSRBench, but exhibits a notable performance drop of 9.9\% on CVBench-3D (see Table~\ref{tab:distance_performance}).

\begin{figure}[t]
    \centering
    \includegraphics[width=\linewidth]{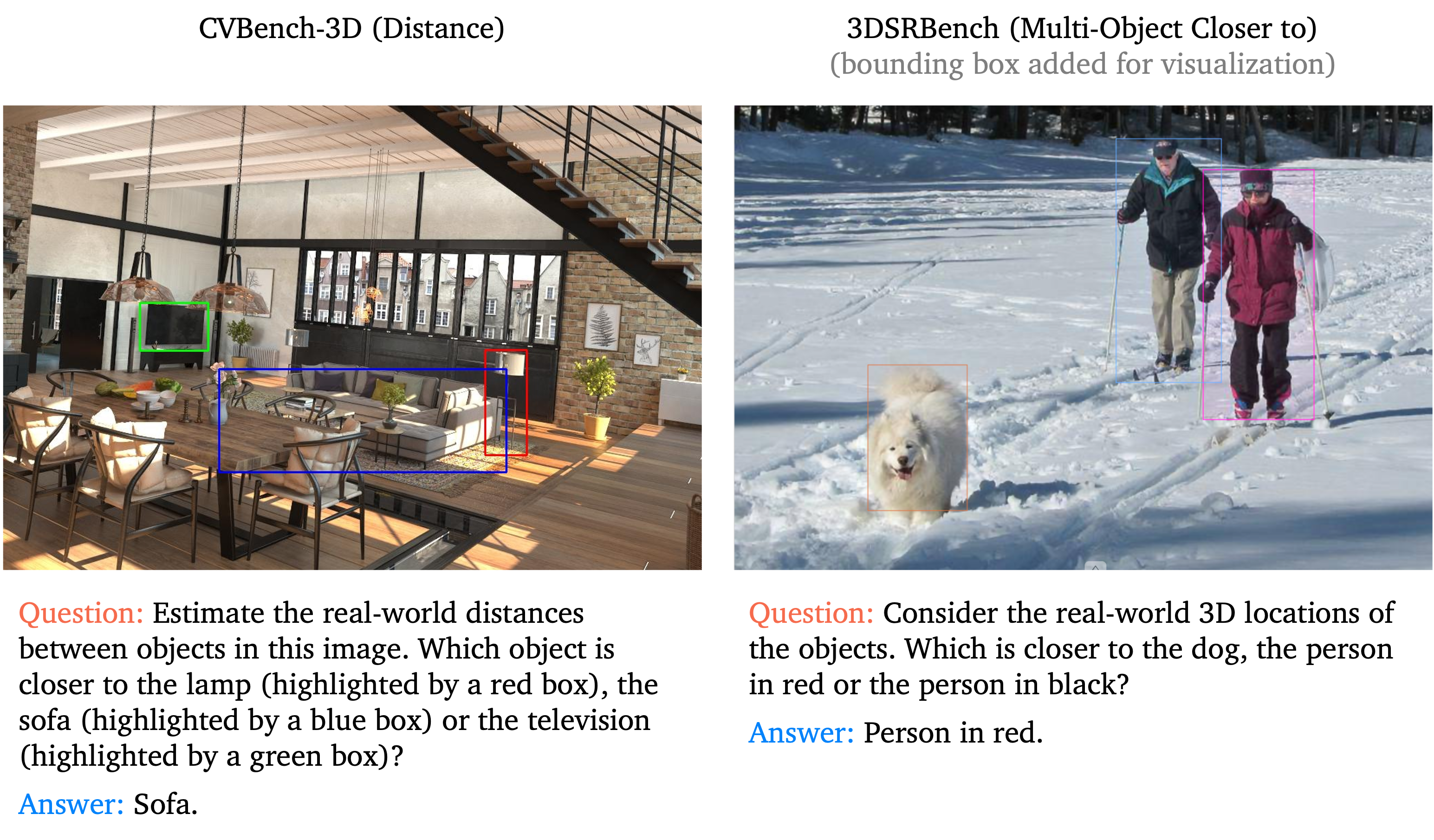}
    \caption{Comparison between (multi-object) distance-related questions in CVBench-3D~\cite{tong2024cambrian} and 3DSRBench~\cite{ma20243dsrbench}.}
    \label{fig:compare_shortcuts}
\end{figure}

\textbf{We attribute this discrepancy to the abundant shortcuts in distance-related questions in CVBench-3D.} From the qualitative examples in Figure~\ref{fig:compare_shortcuts}, the provided 2D bounding boxes in CVBench-3D can be exploited as shortcuts to answer the 3D spatial reasoning question. Rather than reasoning about 3D distances between objects, we can easily derive the correct answer by comparing the 2D distances between the red and blue boxes and between the red and green boxes. Meanwhile, 3DSRBench is a human-collected VQA dataset and manually avoid such spurious correlation, \textit{e.g.}, ``objects closer in 3D space are also closer in 2D image plane''. For the 3DSRBench example in Figure~\ref{fig:compare_shortcuts}, the bounding box of the dog is actually closer to the bounding box of the man in black that farther away in 3D space.

Given the 2D bounding box annotations in CVBench-3D and 3DSRBench, we derive a simple heuristic that attempts to answer distance-related spatial reasoning questions by simply comparing L2 distances between 2D centers of the object bounding boxes. We achieve a 80.2\% accuracy on distance questions in CVBench-3D and 34.3\% in 3DSRBench. \textbf{This demonstrates that baseline models such as Qwen2.5-VL are largely exploiting 2D spatial reasoning as a shortcut to answer complex 3D spatial reasoning questions.} Meanwhile, with our 3D-aware post-training, our \modelname{} adopts explicit 3D representation for various 3D spatial reasoning questions. \textbf{The trade-off between exploiting 2D shortcuts and adopting explicit 3D representations results in slightly lower performance of \modelname{} on test data with abundant spurious correlations, but more importantly, leads to a robust and largely improved performance on a challenging real-world datasets.}

\insightbox{Visual cues or explicit 3D reasoning? LVLMs that exploit 2D reasoning as shortcuts may achieve improved performance on test data with abundant spurious correlations (\textit{e.g.}, distance questions in CVBench-3D). However, they cannot genuinely solve 3D spatial reasoning problems and fall far behind \modelname{} that builds on explicit 3D spatial reasoning when tested on challenging real-world problems in 3DSRBench.}

\section{Open Access}

We release all code, data, and models to support reproducibility and benefit the research community. See links on our \href{https://spatial-reasoner.github.io/}{project page}.

\begin{table}[t]
    \centering
    \footnotesize
    \begin{tabular}{ccc}
        \toprule
         & CVBench3D & 3DSRBench \\
        Method & Distance & multi-object-closer-to \\
        \midrule
        2D Heuristic & 80.2 & 34.3 \\
        Qwen2.5-VL-7B-Instruct~\cite{Qwen2.5-VL} & 83.2 & 49.4 \\
        \modelname & 73.3 & 70.9 \\
        \bottomrule
    \end{tabular}
    \caption{Comparison between Qwen2.5-VL~\cite{Qwen2.5-VL} and \modelname{} on (multi-object) 3D distance-related questions in CVBench-3D~\cite{tong2024cambrian} and 3DSRBench~\cite{ma20243dsrbench}.}
    \label{tab:distance_performance}
\end{table}

\end{document}